\documentclass[conference]{IEEEtran}
\IEEEoverridecommandlockouts

\usepackage{cite}
\usepackage{amsmath,amssymb,amsfonts}
\usepackage{algorithmic}
\usepackage{graphicx}
\usepackage{textcomp}
\usepackage{xcolor}
\usepackage{booktabs}
\def\BibTeX{{\rm B\kern-.05em{\sc i\kern-.025em b}\kern-.08em
    T\kern-.1667em\lower.7ex\hbox{E}\kern-.125emX}}
\begin{document}

\title{LLM-Based Weak Annotation for Cybersecurity Tweet Classification: A Teacher--Student Framework}

\author{
\IEEEauthorblockN{Ferry Afit Kurniawan, Kazushi Okamoto, Koki Karube, Kei Harada, Atsushi Shibata}
\IEEEauthorblockA{Department of Informatics, Graduate School of Informatics and Engineering, The University of Electro-Communications}
\IEEEauthorblockA{1-5-1 Chofugaoka, Chofu, Tokyo 182-8585, Japan}
\IEEEauthorblockA{\{ferry.afit, kazushi, karubekoki, harada, shibata-atsushi\}@uec.ac.jp}
}

\IEEEaftertitletext{%
  \begin{minipage}{\textwidth}
    \footnotesize
    \raggedright
    \copyright\ 2026 IEEE.
    Personal use of this material is permitted.
    Permission from IEEE must be obtained for all other uses, in any current or future media, including reprinting/republishing this material for advertising or promotional purposes, creating new collective works, for resale or redistribution to servers or lists, or reuse of any copyrighted component of this work in other works.
    \par
  \end{minipage}
  \vspace{1em}
}

\maketitle

\begin{abstract}
Cyber threat intelligence (CTI) relies on the timely identification of emerging threats, enabling organizations and security analysts to support situational awareness and decision-making.
Social media platforms such as X (formerly Twitter) have become valuable CTI sources because vulnerabilities and attack activities are often discussed there before appearing in formal reports.
However, the rapid growth of cybersecurity discussions on social media has created more content than domain experts can realistically review and annotate, making labeled data a major bottleneck for developing supervised learning approaches.
While large language models (LLMs) have shown promise as weak annotators in other domains, their effectiveness for cybersecurity tweet classification and downstream training of domain-specific pre-trained language models (PLMs) remains underexplored.
To address this challenge, we propose a Teacher--Student weak supervision framework in which a selected LLM serves as a Teacher to generate weak labels for unlabeled cybersecurity tweets, enabling the training of cybersecurity-specific PLMs without large-scale manual annotation.
Experimental results show that PLMs trained solely on LLM-generated labels consistently achieve F1 scores above 0.92, remaining within 3 percentage points of fully human-supervised models across all evaluated architectures.
These findings suggest that LLM-based weak annotation offers a practical alternative for constructing cybersecurity-specific training data when expert annotation resources are limited.
\end{abstract}


\section{Introduction}
\label{sec:intro}
Cyber threat intelligence (CTI) supports cybersecurity decision-making by providing actionable information about emerging threats, vulnerabilities, and adversarial activity at strategic, operational, and tactical levels~\cite{b1,b2}.
Modern CTI systems combine heterogeneous information sources, knowledge bases, and analytical tools to transform raw threat data into insights that support security analysts and organizational decisions~\cite{b1}.
Among these sources, social media platforms such as X (formerly Twitter) have become valuable channels for collecting timely CTI.
Cybersecurity researchers, vendors, and practitioners often share newly discovered vulnerabilities, malware campaigns, exploit activities, and indicators of compromise (IOCs) on these platforms before the same information appears in formal advisories or public databases~\cite{b3,b4,b5,b6,b7}.
Automated analysis of social media streams can therefore complement traditional threat intelligence pipelines and improve cyber situational awareness.

Despite this potential, cybersecurity-related tweets are noisy, brief, and produced at a volume that far exceeds what experts can realistically review and annotate manually~\cite{b5,b7}.
This creates a bottleneck for supervised learning approaches, which depend on high-quality labeled data for training.
Although numerous models have been proposed for cybersecurity tweet classification and related CTI tasks, their effectiveness depends on carefully curated annotated datasets~\cite{b3,b4,b8,b9, b10, b13}.

Weak supervision offers a promising way to reduce the reliance on exhaustive manual annotation.
Additionally, recent advances in natural language processing (NLP) have demonstrated that large language models (LLMs) can generate weak annotations for training downstream classifiers in a variety of text classification tasks~\cite{b12,b17,b18,b19,b20}.
However, the reliability of LLM-generated weak labels for cybersecurity tweet classification, and their usefulness for training cybersecurity-specific pre-trained language models (PLMs), remain insufficiently explored.

To address this gap, we propose a \textbf{Teacher-Student weak supervision framework} for cybersecurity tweet classification.  
First, multiple candidate LLMs are evaluated as Teacher models.
Next, the most reliable model is selected to annotate a large corpus of unlabeled cybersecurity tweets.
Finally, these weakly labeled data are used to fine-tune cybersecurity-specific PLMs, and their performance is compared under human-only, LLM-only, and hybrid supervision settings.
This design enables us to examine both the annotation quality of LLM Teachers and the downstream value of their weak labels for domain-specific Student models.
Accordingly, this study is guided by three research questions:
\begin{description}
\item[\textbf{RQ1}] Which candidate LLM provides the most reliable weak annotations for cybersecurity tweets?
\item[\textbf{RQ2}] Can LLM-generated weak labels effectively supervise the training of cybersecurity-specific PLMs?
\item[\textbf{RQ3}] How do LLM-only, human-only, and hybrid supervision compare in downstream classification performance?
\end{description}

The main contributions of this work are summarized as follows:
\begin{itemize}
\item We systematically evaluate three LLMs as weak annotators for cybersecurity tweet classification across public benchmark datasets and a newly collected gold-standard dataset.

\item We introduce a scalable Teacher-Student pipeline that uses LLM-generated weak labels to construct a training set for cybersecurity-specific PLMs, reducing the need for extensive manual annotation.

\item Across four cybersecurity-specific PLMs, we show that LLM-only training recovers approximately 98\% of human-only F1 on average, indicating that weak annotation can provide competitive downstream performance under limited annotation resources.
\end{itemize}

\section{Related Work}
\label{sec:related}
\subsection{Cybersecurity Tweet Classification}
Initial work on cybersecurity tweet classification centered on automatically identifying cybersecurity-related content in social media streams.
Dion\'{i}sio et al.~\cite{b3} applied deep neural networks to cyber threat detection on Twitter, while Alves et al.~\cite{b4} investigated large-scale tweet processing for cybersecurity threat awareness.
Subsequent studies explored novelty detection~\cite{b5}, contrastive word embeddings~\cite{b8}, and hybrid IDCNN--BiLSTM architectures~\cite{b9} to improve classification performance.
More recently, transformer-based approaches have become increasingly prominent.
Marinho and Holanda~\cite{b13} applied NLP techniques to automated CTI, whereas Bayer et al.~\cite{b14}, Kebeli and Kwon~\cite{b15}, and You et al.~\cite{b16} explored fine-tuning, self-training, and prompt-based learning to reduce dependence on large labeled datasets while maintaining competitive performance.
These studies demonstrate the effectiveness of modern NLP methods for CTI-oriented tweet classification, but they still primarily focus on model design or task-specific supervision rather than LLM-generated weak labels for cybersecurity-specific PLMs.

\subsection{LLM-Based Weak Supervision}
Recent advances in LLMs have enabled their use as automatic annotators for generating weak labels in NLP tasks.
Gilardi et al.~\cite{b17} reported that ChatGPT can achieve annotation quality comparable to or better than that of human crowd workers across several classification tasks.
He et al. proposed AnnoLLM~\cite{b18}, which improves annotation quality through explanation-guided prompting, while Pangakis and Wolken~\cite{b12} demonstrated that LLM-generated labels can effectively supervise downstream classifiers through knowledge distillation.
Similarly, Bansal and Sharma~\cite{b19} showed that LLM-based annotation can improve model generalization at relatively low labeling cost.
Beyond fully automated annotation, hybrid annotation methods have also attracted increasing attention.
MEGAnno+~\cite{b20} provides an interactive human--LLM collaborative annotation environment, and Wang et al.~\cite{b21} proposed verification mechanisms that combine LLM-generated labels with selective human review to improve reliability and efficiency.

\subsection{Research Gap and Positioning}
Although prior studies have advanced cybersecurity tweet classification and LLM-based weak annotation, the intersection of these two research directions remains largely unexplored.
Existing cybersecurity studies mainly focus on developing classification architectures or reducing supervision through self-training and prompt-based learning, whereas LLM-based annotation has been evaluated predominantly on general-domain NLP tasks.
Consequently, it remains unclear whether LLM-generated weak labels are reliable enough to train cybersecurity-specific PLMs for CTI-oriented tweet classification.
Recent Teacher--Student frameworks have demonstrated the feasibility of training classifiers from LLM-generated labels without manual annotation~\cite{b22}; however, their applicability to cybersecurity-specific PLMs and to supervision regimes that combine human and LLM labels has not been systematically investigated.
This work addresses this gap by evaluating multiple LLMs as Teacher models and assessing whether their weak labels can effectively train cybersecurity-specific PLMs under human-only, LLM-only, and hybrid supervision settings.

Table~\ref{tab:related} summarizes representative prior studies and positions the proposed framework relative to existing research.

\begin{table*}[t]
\centering
\caption{Comparison of representative studies and the proposed work.}

\begin{tabular}{l l c c c c}
\toprule
\textbf{Study} &
\textbf{Primary Focus} &
\textbf{Cybersecurity} &
\textbf{LLM-based} &
\textbf{Trains on} &
\textbf{Human/} \\
&
&
\textbf{Domain} &
\textbf{Weak Labels} &
\textbf{Weak Labels} &
\textbf{LLM/Hybrid} \\
\midrule

Dion\'{i}sio et al.~\cite{b3}
& DNN classification
& \checkmark
& -- & -- & -- \\

Shin et al.~\cite{b8}
& Contrastive embedding
& \checkmark
& -- & -- & -- \\

Bayer et al.~\cite{b14}
& Few-shot learning
& \checkmark
& -- & -- & -- \\

TI-Prompt~\cite{b16}
& Prompt tuning
& \checkmark
& -- & -- & -- \\

CTI-ANN~\cite{b15}
& Self-training annotation
& \checkmark
& -- & \checkmark & -- \\

Pangakis and Wolken~\cite{b12}
& LLM-generated labels
& --
& \checkmark & \checkmark & -- \\

AnnoLLM~\cite{b18}
& LLM annotation framework
& --
& \checkmark & -- & -- \\

MEGAnno+~\cite{b20}
& Human--LLM collaboration
& --
& \checkmark & -- & \checkmark \\

Kuzman and Ljube\v{s}i\'{c}~\cite{b22}
& Teacher--Student learning
& --
& \checkmark & \checkmark & -- \\

\textbf{This Work}
& \textbf{Teacher--Student weak supervision}
& \textbf{\checkmark}
& \textbf{\checkmark}
& \textbf{\checkmark}
& \textbf{\checkmark} \\

\bottomrule
\end{tabular}
\label{tab:related}
\end{table*}

\section{Methodology}
\label{sec:method}
\subsection{Overall Framework}
\begin{figure*}
    \centering
    \includegraphics[width=\linewidth]{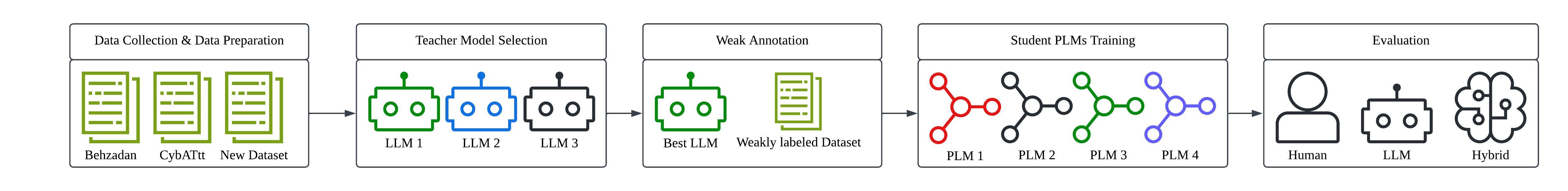}
    \caption{Teacher--Student framework}
    \label{fig:framework}
\end{figure*}

The framework consists of five sequential stages, as illustrated in Fig.~\ref{fig:framework}.
First, data sources are prepared by combining public benchmark datasets with a newly collected cybersecurity tweet corpus, including a manually annotated subset for evaluation and supervised baselines.
Second, multiple candidate LLMs are evaluated as potential Teacher models, and the most reliable model is selected based on annotation performance.
Third, the selected Teacher assigns weak labels to previously unlabeled cybersecurity tweets, producing a large weakly labeled corpus.
Fourth, this corpus is used to fine-tune multiple cybersecurity-specific PLMs, which serve as Student models.
Finally, the resulting classifiers are evaluated and compared across human-only, LLM-only, and hybrid supervision settings.

\subsection{Datasets}
\subsubsection{Data Collection}
To construct a new cybersecurity tweet corpus, we collected tweets from 112 cybersecurity-related X accounts between November 2024 and January 2025.
The selected accounts represent diverse and reputable sources of CTI, including  49 cybersecurity researchers, 39 security vendors, 7 cybersecurity organizations and agencies, and 17 cybersecurity news outlets.
A total of 17,581 tweets were collected.
After removing non-English tweets and duplicates and normalizing the text, 13,451 unique English-language tweets remained.
To build a gold-standard benchmark, we selected a stratified sample of 2,000 tweets for manual annotation.
Each tweet was independently labeled by two annotators with expertise in cybersecurity, following predefined annotation guidelines.
Inter-annotator agreement was measured using Cohen's Kappa ($\kappa = 0.89$), which indicates high agreement.
To ensure high label quality, only tweets for which both annotators assigned the same label were retained.
As a result, 103 tweets with annotation disagreements were excluded, yielding a final gold-standard dataset of 1,897 tweets.
\subsubsection{Data Splitting}
Among the annotated tweets, 400 were reserved as an independent held-out test set and were not used  for Teacher selection, weak annotation, or Student model training.
The remaining 1,497 manually labeled tweets were used in the human-only and hybrid supervision settings.
For Teacher selection, 400 tweets were sampled from this 1,497-tweet pool.
The 11,451 tweets outside the annotated subset remained unlabeled and were annotated automatically by the selected Teacher model.
These weakly labeled tweets were subsequently used to train the Student models.

\subsubsection{Public Benchmark Datasets}
In addition to the newly collected dataset, we used the publicly available Behzadan~\cite{b23} and CybAttT~\cite{b24} datasets during Teacher model selection.
Evaluating the candidate Teachers on multiple datasets helps assess performance across different data distributions and reduces the risk of selecting a model that performs well on only a single benchmark.
\subsubsection{Label Harmonization}
The three datasets used in this study were originally annotated using different label schemas.
To enable a consistent evaluation of Teacher models and downstream Student training, all datasets were converted into a unified binary classification task consisting of \textit{Relevant} and \textit{Irrelevant} classes.
Tweets were annotated as \textit{Relevant} if they contained concrete cyber threat information, such as CVEs and other vulnerabilities, exploits, malware or ransomware, active attacks, security incidents, IOCs, or attributed threat actors and campaigns.
Tweets that did not contain concrete threat information were labeled as \textit{Irrelevant}.
The newly collected dataset was already annotated using the binary labels \textit{Relevant} and \textit{Irrelevant} and therefore required no label harmonization.
For the CybAttT dataset, both \textit{Normal News} and \textit{High Risk News} were mapped to the \textit{Relevant} class, whereas \textit{Not News} was mapped to the \textit{Irrelevant} class.
For the Behzadan dataset, the \textit{Threat} category was mapped to the \textit{Relevant} class, while the \textit{Business} and \textit{Irrelevant} categories were merged into the \textit{Irrelevant} class.
Table~\ref{tab:label_mapping} summarizes the label harmonization process across the three datasets. 

\begin{table}[t]
\centering
\setlength{\tabcolsep}{2.5pt}
\caption{Label harmonization across datasets.}
\label{tab:label_mapping}
\begin{tabular}{llll}
\toprule
\textbf{Dataset} & \textbf{Original Label (s)} & \textbf{Binary Label} & \textbf{Count} \\
\midrule

Our Data (Gold) &
Relevant &
Relevant &
1,169 \\

Our Data (Gold) &
Irrelevant &
Irrelevant &
728 \\

CybAttT &
Normal News + High Risk News &
Relevant &
4,840 \\

CybAttT &
Not News &
Irrelevant &
31,231 \\

Behzadan &
Threat &
Relevant &
8,280 \\

Behzadan &
Business + Irrelevant &
Irrelevant &
8,929 \\

\bottomrule
\end{tabular}
\end{table}

\subsection{Teacher Model Selection}
We evaluated three candidate LLMs: GPT-4.1-nano, GPT-4o-mini, and Llama-3.3-70B Instruct.
The evaluation was conducted using manually labeled samples drawn from the three benchmark datasets described previously.
Specifically, 150 Relevant and 150 Irrelevant tweets were sampled from each of Behzadan and CybAttT, and 400 tweets were sampled from the gold-data pool described above, resulting in 1,000 evaluation samples.
All candidate models were evaluated under the same zero-shot prompting setting.
For each tweet, the model was asked to determine whether the content was \textit{Relevant} or \textit{Irrelevant} to cybersecurity.
The predicted labels were then compared with the human-annotated ground truth. The complete prompt template is shown in Fig.~\ref{fig:prompt}.

\begin{figure}[t]
\centering
\fbox{\parbox{0.92\columnwidth}{\small
\textbf{Zero-shot prompt (universal across datasets, $T = 0$):}\\[2pt]
\textit{Role:} Senior Cyber Threat Intelligence Analyst.\\
\textit{Task:} Classify the tweet as Relevant (1) or Irrelevant (0).

A tweet is Relevant (1) if it contains CONCRETE information about 
a cyber threat -- a specific CVE or vulnerability, a named exploit, 
a named malware family, an active attack or security incident with 
a named victim or target, an indicator of compromise (IP, hash, 
domain, file path), or an attributed threat actor or campaign.

A tweet is Irrelevant (0) if it contains NO concrete threat 
instance -- including general cybersecurity discussion, security 
news commentary, marketing, job postings, opinions, educational 
content, policy discussions, software releases, or statistical 
trends about cybersecurity without naming a specific incident.

Operative test: does this tweet reference a SPECIFIC, NAMED 
threat instance that a CTI analyst could act on? If not, label 0.

When genuinely uncertain, default to 0.
\\
\\
\textit{Constraint:} Output ONLY the digit 0 or 1.}}
\caption{Universal zero-shot annotation prompt, used identically across
all Teacher LLM candidates and all three datasets.}
\label{fig:prompt}
\end{figure}

\subsection{Weak Annotation}
The selected Teacher model was applied to 11,451 unlabeled cybersecurity tweets to generate weak labels for downstream model training.
The Teacher produced 7,275 \textit{Relevant} (63.5\%) and 4,176 \textit{Irrelevant} labels (36.5\%).
The resulting weakly labeled corpus served as the training data for the LLM-only setting and as an additional supervision source in the hybrid setting.
Thus, the downstream experiments evaluate the utility of the raw Teacher-generated labels without manual correction.

\subsection{Student PLM Training}
Four cybersecurity-specific PLMs were selected as Student models: CTI-BERT~\cite{b11}, CySecBERT~\cite{b25}, SecureBERT-2.0~\cite{b26}, and SecRoBERTa~\cite{b27}.
These models were chosen because they were pre-trained on cybersecurity-related corpora and represent diverse architectures and training strategies within the cybersecurity NLP domain.
Using multiple Student models reduces the likelihood that the findings are tied to a particular architecture and provides a broader assessment of the proposed framework.

\begin{itemize}
    \item \textit{Human-only}: trained using the 1,497 manually annotated tweets.
    \item \textit{LLM-only}: trained using the 11,451 weak labels generated by the selected Teacher model.
    \item \textit{Hybrid}: trained using the combination of 1,497 manually annotated tweets and 11,451 LLM-generated labels.
\end{itemize}

\subsection{Evaluation Procedure}
Teacher model performance was measured on the 1,000-tweet evaluation set described in Sec.~III-C using accuracy, precision, recall, and F1-score.
The model achieving the strongest overall F1 across the three datasets is selected as the Teacher and used for all subsequent weak annotation.
Student PLM performance was measured on the 400-tweet held-out test set, which was set aside before any training and never used for model selection.
For comparisons among supervision regimes, we applied an exact permutation Friedman test to the five paired seed-level F1-scores for each PLM.
Pairwise Wilcoxon signed-rank comparisons with Holm correction were reported descriptively because five paired seeds limit the attainable exact $p$-values.

\section{Experiment and Results}
\label{sec:experiment}
\subsection{Teacher Model Selection}
Table~\ref{tab:teacher} reports per-dataset performance.
On CybAttT, all three models achieved F1 scores above 0.82, while F1 scores decreased for all three candidates on Behzadan.
This decrease was mainly associated with lower recall, suggesting that some Behzadan tweets mapped as Relevant after label harmonization may have fallen outside the concrete cyber threat information criterion used in our prompt.
On the gold-standard evaluation sample, Llama-3.3-70B achieved the highest F1 score (0.9225), closely followed by GPT-4.1-nano (0.9137).

\begin{table}[t]
\caption{Zero-Shot LLM Annotation Quality Per Dataset}
\label{tab:teacher}
\centering
\setlength{\tabcolsep}{2.5pt}
\begin{tabular}{llcccc}
\toprule
\textbf{Dataset} & \textbf{Model}
  & \textbf{Accuracy} & \textbf{Precision} & \textbf{Recall}
  & \textbf{F1-score} \\
\midrule
Behzadan
  & GPT-4.1-nano  & 0.7000 & 0.8125 & 0.5200 & 0.6341 \\
  & GPT-4o-mini   & 0.6567 & 0.8310 & 0.3933 & 0.5339 \\
  & Llama-3.3-70B & 0.7500 & 0.8261 & 0.6333 & \textbf{0.7170} \\
CybAttT
  & GPT-4.1-nano  & 0.8533 & 0.8193 & 0.9067 & \textbf{0.8608} \\
  & GPT-4o-mini   & 0.8267 & 0.8500 & 0.7933 & 0.8207 \\
  & Llama-3.3-70B & 0.7967 & 0.7214 & 0.9667 & 0.8262 \\
Our Data
  & GPT-4.1-nano  & 0.8975 & 0.9518 & 0.8785 & 0.9137 \\
  & GPT-4o-mini   & 0.8500 & 0.9652 & 0.7854 & 0.8661 \\
  & Llama-3.3-70B & 0.9025 & 0.9063 & 0.9393 & \textbf{0.9225} \\
\bottomrule
\end{tabular}
\end{table}
\begin{table*}[t]
\caption{PLM Classification Performance}
\label{tab:plm}
\centering
\begin{tabular}{llcccc}
\toprule
\textbf{Model} & \textbf{Regime}
  & \textbf{Accuracy} & \textbf{Precision} & \textbf{Recall}
  & \textbf{F1-score} \\
\midrule
CTI-BERT
  & Human-only 
  & $\mathbf{0.9445 \pm 0.0062}$ 
  & $\mathbf{0.9652 \pm 0.0097}$
  & $\mathbf{0.9439 \pm 0.0150}$ 
  & $\mathbf{0.9543 \pm 0.0054}$ \\
  & LLM-only   
  & $0.9145 \pm 0.0033$ 
  & $0.9220 \pm 0.0072$ 
  & $0.9407 \pm 0.0068$ 
  & $0.9312 \pm 0.0025$          \\
  & Hybrid     
  & $0.9155 \pm 0.0041$ 
  & $0.9262 \pm 0.0069$
  & $0.9374 \pm 0.0140$ 
  & $0.9317 \pm 0.0040$          \\
CySecBERT
  & Human-only 
  & $\mathbf{0.9410 \pm 0.0045}$ 
  & $\mathbf{0.9646 \pm 0.0163}$ 
  & $0.9390 \pm 0.0228$ 
  & $\mathbf{0.9513 \pm 0.0045}$ \\
  & LLM-only   
  & $0.9180 \pm 0.0086$ 
  & $0.9328 \pm 0.0098$ 
  & $0.9342 \pm 0.0183$ 
  & $0.9333 \pm 0.0075$          \\
  & Hybrid     
  & $0.9210 \pm 0.0058$ 
  & $0.9282 \pm 0.0064$ 
  & $\mathbf{0.9447 \pm 0.0154}$ 
  & $0.9363 \pm 0.0052$          \\
SecRoBERTa
  & Human-only 
  & $0.9205 \pm 0.0105$ 
  & $0.9340 \pm 0.0166$ 
  & $0.9374 \pm 0.0231$ 
  & $0.9354 \pm 0.0090$          \\
  & LLM-only   
  & $0.9065 \pm 0.0058$ 
  & $0.9192 \pm 0.0133$ 
  & $0.9301 \pm 0.0202$ 
  & $0.9244 \pm 0.0054$          \\
  & Hybrid     
  & $\mathbf{0.9250 \pm 0.0068}$ 
  & $\mathbf{0.9364 \pm 0.0092}$ 
  & $\mathbf{0.9423 \pm 0.0171}$ 
  & $\mathbf{0.9392 \pm 0.0061}$ \\
SecureBERT-2.0
  & Human-only 
  & $\mathbf{0.9370 \pm 0.0045}$ 
  & $\mathbf{0.9744 \pm 0.0098}$ 
  & $0.9220 \pm 0.0123$ 
  & $\mathbf{0.9474 \pm 0.0040}$ \\
  & LLM-only   
  & $0.9075 \pm 0.0105$ 
  & $0.9132 \pm 0.0133$ 
  & $0.9390 \pm 0.0119$ 
  & $0.9259 \pm 0.0083$          \\
  & Hybrid     
  & $0.9325 \pm 0.0047$ 
  & $0.9405 \pm 0.0026$ 
  & $\mathbf{0.9504 \pm 0.0105}$ 
  & $0.9454 \pm 0.0041$          \\
\bottomrule
\end{tabular}
\end{table*}
\begin{table}[t]
\caption{Size-Matched LLM-only Ablation }
\label{tab:exp1}
\centering
\setlength{\tabcolsep}{2.5pt}
\begin{tabular}{lccc}
\toprule
\textbf{Model}
  & \textbf{LLM-matched}
  & \textbf{LLM-full}
  & \textbf{Human-only} \\
\midrule
CTI-BERT        
& $0.9240 \pm 0.0062$ 
& $0.9312 \pm 0.0025$ 
& $0.9543 \pm 0.0054$\\
CySecBERT      
& $0.9220 \pm 0.0062$ 
& $0.9333 \pm 0.0075$ 
& $0.9513 \pm 0.0045$\\
SecRoBERTa      
& $0.9020 \pm 0.0134$ 
& $0.9244 \pm 0.0054$ 
& $0.9354 \pm 0.0090$ \\
SecureBERT-2.0  
& $0.9171 \pm 0.0062$ 
& $0.9259 \pm 0.0083$ 
& $0.9474 \pm 0.0040$\\
\bottomrule
\end{tabular}
\end{table}

Across the full 1,000-tweet evaluation set, Llama-3.3-70B achieved the highest F1 score (0.8436), followed by GPT-4.1-nano (0.8312) and GPT-4o-mini (0.7758).
McNemar's test confirmed that both top-performing models significantly outperformed GPT-4o-mini ($\chi^{2} \geq 9.51$, $p < 0.01$ for both comparisons).
Although the difference between Llama-3.3-70B and GPT-4.1-nano was not significant ($\chi^{2} = 0.01$, $p = 0.92$), Llama-3.3-70B was selected as the Teacher based on its highest F1 score.
It also achieved higher recall on Our Data (0.9393 vs. 0.8785), which is desirable for CTI pre-filtering.
However, GPT-4.1-nano incurred a lower annotation cost than Llama-3.3-70B  (\$0.0100 versus \$0.0158) for 400 tweets, indicating a cost--performance trade-off.

\subsection{Student PLM Classification}
Table~\ref{tab:plm} reports PLM performance on the 400-tweet held-out test set.
Each training configuration was run five times with different random seeds; results are reported as mean~$\pm$~standard deviation across the five runs.
Across all four PLMs, LLM-only training recovered approximately 98\% of the human-only F1 score on average (range 97.6\%--98.8\%).
The largest mean F1 gap was observed for CTI-BERT (0.9543 versus 0.9312).
Hybrid training achieved a higher mean F1 than LLM-only training for all PLMs and achieved the highest mean F1 for SecRoBERTa (0.9392).

Exact permutation Friedman tests indicated significant overall differences among the supervision regimes for all PLMs (Table~IV: $Q=6.4$--$8.4$, $p=0.008$--$0.039$; Table~V: $Q=8.4$--$10.0$, $p<0.009$).
With five paired seeds, the smallest attainable two-sided exact Wilcoxon $p$-value is 0.0625.
No Holm-adjusted pairwise comparison can therefore reach the 0.05 threshold.
Pairwise differences are consequently interpreted descriptively based on mean F1.

\subsection{Size-matched LLM-only Ablation}
To disentangle the contribution of label \textit{quantity} from label \textit{quality}, each PLM was also trained on a size-matched LLM-only subset of $n = 1{,}497$ tweets.
This subset matched the size of the human-only pool and was drawn from the full LLM-only corpus using stratified sampling.
The subset was drawn once and held fixed across seeds.
Thus, the reported variability does not reflect variation in subset selection.
Table~\ref{tab:exp1} compares these results with those obtained from full LLM-only and human-only training.
When restricted to the same number of labels ($n = 1{,}497$), human-only training yielded 2.9--3.3 percentage points higher mean F1 than size-matched LLM-only training across all four models, suggesting that manual labels provided a stronger supervision signal at equal scale.
The full LLM-only corpus (11,451 tweets; 7.6$\times$ the human-only pool) reduced this mean F1 gap to 1.1--2.3 percentage points.
The complete workflow, comprising evaluation of three candidate Teachers on the 1,000-tweet evaluation set and weak annotation of the 11,451 unlabeled tweets, incurred total API charges of less than US\$0.70.

\subsection{Discussion}
\subsubsection{RQ1: Which candidate LLM provides the most reliable weak annotations for cybersecurity tweets?}
Llama-3.3-70B was selected as the Teacher model because it achieved the highest overall F1-score and the highest recall on the gold-standard dataset.
Although its F1-score was not significantly different from that of GPT-4.1-nano ($p = 0.92$), its higher recall is particularly desirable for CTI pipelines, where a relevance classifier serves as the initial filtering stage and missing a genuine threat is generally more costly than flagging additional false positives for review.

\subsubsection{RQ2: Can LLM-generated weak labels effectively supervise the training of cybersecurity-specific PLMs?}
At matched training-set size, human-only training yielded higher mean F1 across all PLMs.
The gap narrowed when the full weakly labeled corpus was used, suggesting that the larger training set contributed to the recovery.
LLM-generated weak labels therefore provide an effective supervision signal, with scale rather than per-label accuracy driving the improvement.

\subsubsection{RQ3: How do LLM-only, human-only, and hybrid supervision compare in downstream classification performance?}
The three supervision settings revealed a trade-off between annotation quality and training data quantity.
Human-only supervision yielded the highest mean F1 for most PLMs.
Hybrid supervision showed a higher mean F1 than LLM-only training for all PLMs but rarely surpassed human-only training, suggesting that additional weak labels may also introduce label noise.
Because the Teacher assigns class labels rather than generating text, this noise takes the form of unsupported label assignments rather than fabricated content.

\subsection{Limitations}
This study has several limitations that provide opportunities for future work.
First, the proposed framework focuses on binary classification, distinguishing \textit{Relevant} from \textit{Irrelevant}.
This design was chosen because relevance filtering is typically the first stage of a CTI pipeline before more specific analyses, such as topic modeling or event detection.
Second, only English-language tweets were included in this study.
Evaluating the framework on multilingual datasets would provide broader coverage, since threat intelligence is also shared in many other languages.
Finally, the newly collected dataset was constructed from 112 curated cybersecurity-related X accounts to reduce noise and improve the efficiency of dataset construction.
However, it may not fully capture cybersecurity discussions from newly emerging sources.
Future research should explore broader collection strategies while maintaining data quality.

\section{Conclusion}
\label{sec:conclusion}
In this study, we proposed and evaluated a Teacher-Student weak supervision framework for cybersecurity tweet classification, in which an LLM serves as the Teacher to generate weak labels for training domain-specific PLMs.
The experimental results showed that Llama-3.3-70B was the most reliable Teacher model, while Student models trained solely on LLM-generated labels achieved performance close to that of fully supervised models.
Additional experiments suggested that scaling training data can partially offset lower label quality.
These findings demonstrate that LLM-based weak annotation offers a practical approach to reducing manual annotation effort while maintaining competitive downstream performance.
For cybersecurity applications, where expert annotation is both expensive and time-consuming, the proposed framework provides an effective alternative for constructing large-scale training datasets with limited human involvement.
Future work should extend this framework by integrating topic modeling and event detection to construct structured CTI events from weakly labeled social media streams.

\end{document}